# Deep Learning based CNN Model for Classification and Detection of Individuals Wearing Face Mask

Dr. R.Chinnaiyan[a], Dr. Iyyappan. M[b], Al Raiyan Shariff A[c], Kondaveeti Sai[d], Mallikarjunaiah B M[e], P Bharath[f]

*vijayachinns@gmail.com [a], iyyappan5mtech@gmail.com [b], shalraiyanbtech17@ced.alliance.edu.in [c], sakondaveetibtech17@ced.alliance.edu.in [d], bmmallikaarjunabtech17@ced.alliance.edu.in [e], pbharathbtech17@ced.alliance.edu.in [f]*

[a*] *Professor, Department of Computer Science and Engineering, Alliance University, Bengaluru, India – 562106.*

[b*] *Associate Professor, Department of Computer Science and Engineering, Alliance University, Bengaluru, India – 562106.*

[c,d,e,f] *Department of Computer Science and Engineering, Alliance University, Bengaluru, India – 562106.*

**Abstract**

In response to the global COVID-19 pandemic, there has been a critical demand for protective measures, with face masks emerging as a primary safeguard. The approach involves a two-fold strategy: first, recognizing the presence of a face by detecting faces, and second, identifying masks on those faces. This project utilizes deep learning to create a model that can detect face masks in real-time streaming video as well as images. Face detection, a facet of object detection, finds applications in diverse fields such as security, biometrics, and law enforcement. Various detector systems worldwide have been developed and implemented, with convolutional neural networks chosen for their superior performance accuracy and speed in object detection. Experimental results attest to the model's excellent accuracy on test data. The primary focus of this research is to enhance security, particularly in sensitive areas. The research paper proposes a rapid image pre-processing method with masks centred on faces. Employing feature extraction and Convolutional Neural Network, the system classifies and detects individuals wearing masks. The research unfolds in three stages: image pre-processing, image cropping, and image classification, collectively contributing to the identification of masked faces. Continuous surveillance through webcams or CCTV cameras ensures constant monitoring, triggering a security alert if a person is detected without a mask.

* Corresponding author.

*E-mail address:* vijayachinns@gmail.com

# 1. Introduction

Deep learning, a subset of machine learning, relies entirely on artificial neural networks. Given that these neural networks emulate the functioning of the human brain, deep learning essentially mimics the cognitive processes of the human mind. Positioned within the broader spectrum of machine learning methods, deep learning employs artificial neural networks with representation learning. Learning within this framework can take the form of supervised, semi-supervised, or unsupervised methods. The essence of deep learning lies in utilizing intricate architectures that amalgamate various non-linear transformations to model complex data. At its core, deep learning employs neural networks, which are then integrated to construct deep neural networks. These methodologies have facilitated notable advancements in domains such as sound and image processing, encompassing applications like facial recognition, speech recognition, computer vision, automated language processing, and text classification (e.g., spam recognition). The potential applications of deep learning are diverse and extensive.

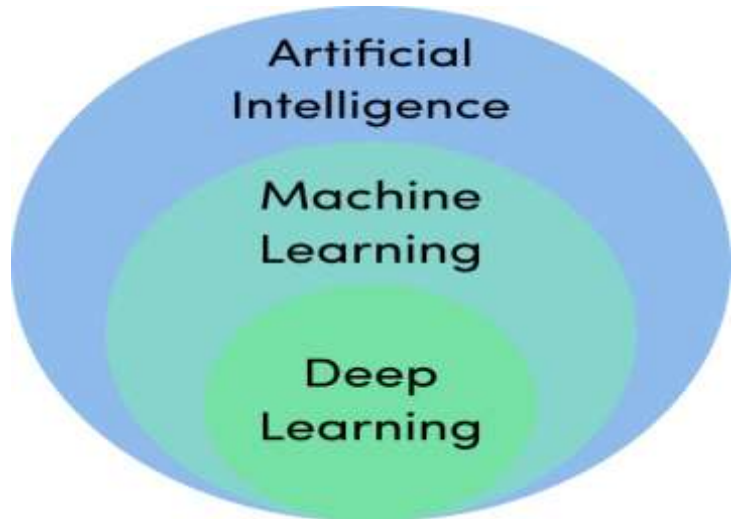

Fig. 1. AI/ML model

**1.1 Face Mask Recognition:** Modern facial recognition software analyses characteristics near the eyes, nose, mouth, and ears to identify an individual based on a supplied image, whether voluntarily provided or sourced from a criminal database. The use of masks poses a challenge to this recognition process, a hurdle that various systems have already grappled with, some successfully addressing. As an illustration, Apple's Face ID, designed for users to unlock their iPhones through facial recognition, recently introduced a system update. This update enables the software to effectively discern when a person is wearing a mask, swiftly acknowledging the covered mouth and nose, and prompting the user to enter their passcode instead of requiring them to remove their face covering.

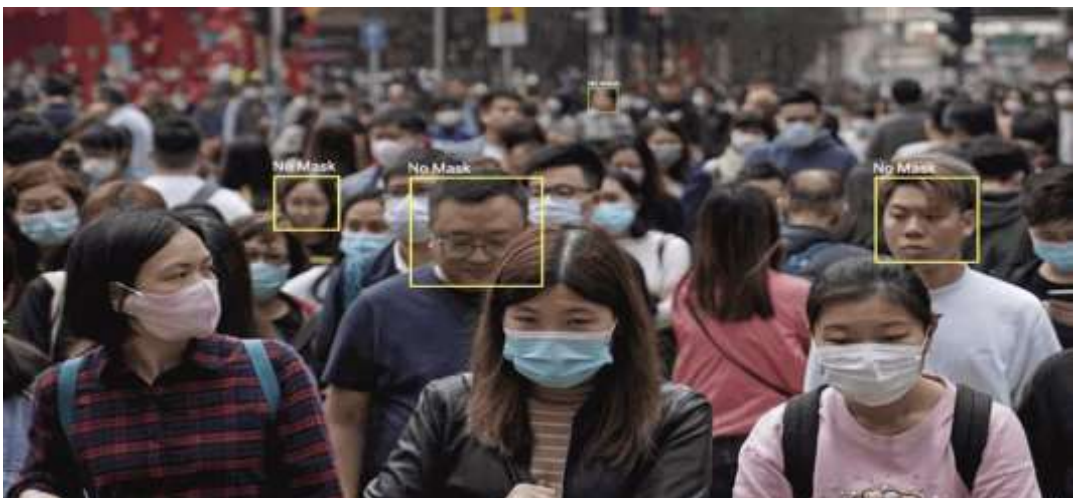

Fig. 2. Face mask recognition

According to developers, mask recognition software theoretically sidesteps privacy concerns since the programs do not identify individuals. These software systems undergo training using two sets of images: one to instruct the algorithm on face recognition ("face detection") and another to instruct it on recognizing masks on faces ("mask recognition"). Importantly, the machine learning algorithm doesn't identify faces in a manner that establishes a link to a specific person. This is because it does not utilize a training set containing examples of faces tied to specific identities.

## 2. Literature Review

Face identification models were previously built utilizing edge, line, and center near features, and patterns were recognized from those features. These methods are used to find binary patterns on a local scale. These algorithms are particularly successful for dealing with grayscale images and involve relatively little computation work [1-2]. AdaBoost is a regression-based classifier that will fit a regression function to the original data set and change wait times during backpropagation to maximize the results. A real-time object model for detecting various object classes was proposed by the Viola-Jones Detector. It analyzes any image having an edge, line, and four rectangular characteristics using a base window size of 24 by 24. Convolutions are like harr-like features in that they determine if a particular feature is present in the image or not [3]. Even though this model performs poorly when images are oriented differently, it fails to function when image brightness varies. Classification issues are the primary application for convolution networks. Different CNN architectures exist, including the VGG-16. This architecture consists of three convolution layers with a max pool, three convolution layers with a max pool, three fully connected layers, and two convolution layers with an input size of 224 kernel (64, 3x3). Then, there are two more convolution layers that follow the max pool. Ultimately, softmax FC When compared to AlexNet, this architecture functions well [4]. Utilizing a fundamentally inception-based architecture, Google Net reduces the number of parameters by building small convolution layers. With about 22 layers that include max-pooling and convolution, among other features, it can function well over AlexNet, reducing its 60 million features to just 4 million [5]. A deep neural network with 152 layers—eight times more than a VGGnet with the least amount of complexity—was proposed by K. Li, G. Ding, and H. Wang. This network uses residual learning to train the models further. Using the COCO data set, this method produced comparatively better item recognition results [6-7]. To performing cardiac ventricular segmentation, X. Fu and H. Qu proposed UNet and SEnet. According to this model, traits that are more valuable are given higher weights, whereas features that are less significant are given less weights. One of the key areas of research that is receiving a lot of interest these days is human posture detection. The author of this research suggested a model that uses a person's stance to identify traditional dancing. They employed CNN to accomplish this, training a model to recognize different traditional dance steps and successfully detecting traditional dancing [8]. There are several methods available for picture analysis, and object detection has grown in importance in the field of image inquiry. The author of this research presented a wavelet-based neural network for learning and feature extraction that is effective in object detection [9]. In order to achieve sine language identification, a CNN model that can recognize signs in real-world videos was trained. It is also highly helpful in autonomous vehicles. It even helps with computer translation in sign language [10]. The planned medical image processing by Lakshmi Ramani Tumuluru was carried out. In this research, instead of utilizing 2D segmentation for tumor detection, we employed 3D segmentation, which they trained using FCN on images of human brains to identify tumors quite effectively [11].

## 3. Design of Proposed System

Gathering information is a crucial first step in creating any real-time detection model, and we used the MaskedFace-Net dataset to build our face mask recognition model. The collection, which comes with 133,783 photos of faces of people wearing masks appropriately or improperly, is taken from Flickr-Faces-HQ (FFHQ). The use of face masks has become a proactive strategy to stop the spread of COVID-19, which calls for effective recognition mechanisms to track adherence in authorized locations. To do this, deep learning models must be trained on a sizable dataset of masked faces in order to distinguish between people who are wearing masks and those who are not. Although there are several sizable datasets of masked faces in the literature, no comprehensive dataset is currently available to evaluate if the masked faces that have been found are adequately worn. Campaigns encouraging appropriate mask-wearing practices are in place because improper mask usage is common owing to a variety of variables, including bad habits, behaviors, or vulnerabilities (e.g., in youngsters or the elderly). To tackle this, we present three different masked face detection datasets in our work: the Correctly Masked Face Dataset (CMFD), the Incorrectly Masked Face Dataset (IMFD), and a combined dataset for all-encompassing masked face recognition (MaskedFace-Net). These datasets have two purposes in mind when it comes to providing a realistic picture of masked faces: to recognize people wearing or without wearing face masks. to recognize faces wearing masks—whether appropriately or incorrectly—in crowds or at airport gateways. As far as we are aware, no sizable dataset of masked faces is available yet that provides this level of precise categorization to enable thorough examination of mask wear. This work also presents a globally applied mask-to-face deformable model, which allows for the creation of different masked face images, especially those with kinds of masks.

**3.1 Image Data Preprocessing:** Information Pre-processing is the term used to describe all the operations done on unprocessed data prior to supplying it to an algorithm for machine learning or deep learning. For example, poor classification performance can arise from trying to train a convolutional neural network directly on raw images. Pre-processing plays a critical role in both improving accuracy and accelerating learning. Neural networks with multiple hidden layers—often hundreds in modern, state-of-the-art models—are used in deep learning, which necessitates large amounts of training data. In perceptual tasks including voice, language processing, and vision, these models have demonstrated remarkable efficacy in obtaining accuracy close to that of humans.

## 4. Design of Proposed System Model

In the Data Collection part, we used freely available open-source Face-Net Dataset. Because of the face net data set they used the existing set of without mask images and converted them as masked images by adding a mask to the face. So, the accuracy and prediction will be more accurate rather than some random masked and without masked images. By using the Collected data set we resize the image to a uniform image format and converted all images to array format and after that we trained our model using MobileNetV2 Architecture in CNN.

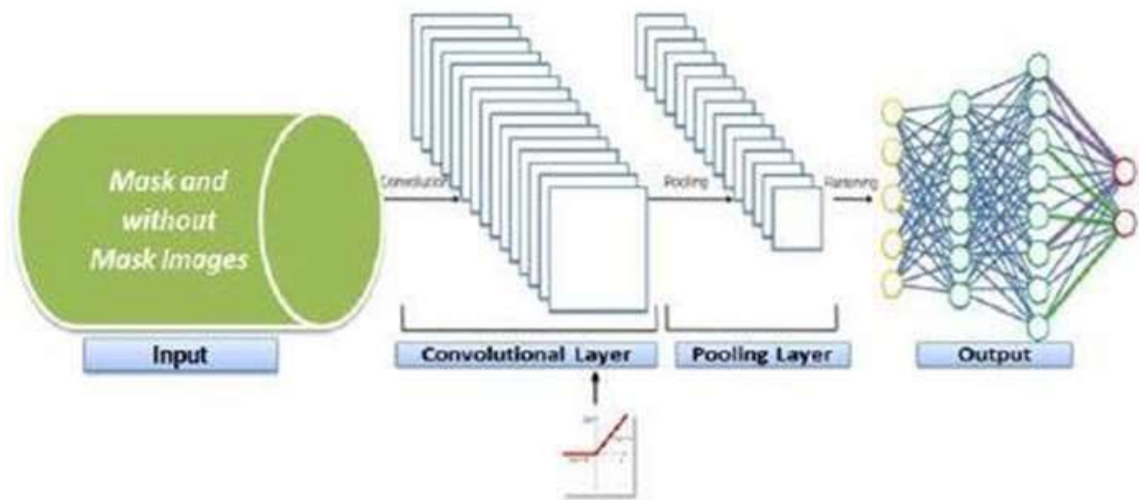

Fig. 3. CNN Model

A convolutional neural network architecture created especially for mobile devices is called MobileNetV2. Remaining connections between bottleneck levels are incorporated, embracing an inverted residual structure. Lightweight depth-wise convolutions are used by the intermediate expansion layer to filter features, adding non-linearity. A first fully convolutional layer with 32 filters makes up the general architecture of MobileNetV2, which is followed by 19 residual bottleneck layers. MobileNetV2, which adds the linear bottleneck layer and inverted residual, improves accuracy and performance for embedded and mobile vision applications. The depth-wise separable convolutions developed in MobileNetV1, which served as the network's basis for MobileNetV2, are expanded upon by this innovative layer. Semantic segmentation, object categorization, and detection are just a few of the tasks that can be customized for this network. And after getting the training Loss and Accuracy curves, we can even get the Model Evaluation details showing the details of precision and accuracy of the system by comparing many parameters. After all the process at the end we take the live video input and read video by frame by frame and capture the photo and resize the image and after that OpenCV will crop the face part and call preprocessing function in that it will predict the image (with mask or with-out mask) and shows the output on the screen.

# 5. Implementation and Result Analysis

For our implementation, we opted for the Python language, renowned for its ease of use and widespread adoption in AI/ML development. Utilizing Jupyter Notebook as our Integrated Development Environment (IDE), we harnessed various libraries and frameworks, including Matplotlib, NumPy, and TensorFlow.

```python
import numpy as np
import matplotlib.pyplot as plt
import os
import cv2
DATADIR = "Dataset/"
#traiing_dataset
categories = ["With_Face_Mask","Without_Mask"]
#list of classes [This will read the datasets present in both with mask and without mask folder and converted them to RGB format because the image original will be in the BGR format]

for cat in categories:
    path = os.path.join(DATADIR,cat)
    for img in os.listdir(path):
        img_array = cv2.imread(os.path.join(path,img))#,cv2.IMREAD_GRAYSCALE)
        plt.imshow(cv2.cvtColor(img_array, cv2.COLOR_BGR2RGB))
        plt.show()
        break
    IMG_SIZE =224
#size must be 224*224

new_array = cv2.resize(img_array,(IMG_SIZE,IMG_SIZE))
plt.imshow(cv2.cvtColor(new_array, cv2.COLOR_BGR2RGB))
plt.show()
```

[creating an array as training data and storing all images in this array with 2 different parameters 0 and 1 for with mask and without mask resp.]

```python
training_data = [] #data
def create_training_data():
    for cat in categories:
        path = os.path.join(DATADIR,cat)
        class_num = categories.index(cat) #0,1 lables
        for img in os.listdir(path):
            try:
                img_array = cv2.imread(os.path.join(path,img))
                new_array = cv2.resize(img_array,(IMG_SIZE,IMG_SIZE))
                training_data.append([new_array,class_num])
            except Exception as e: pass

create_training_data()
```

[Shuffling all the images so we have some random data it will help at the time of data preprocessing, so the model won't be trained for the same type of images while training]

```python
import random
random.shuffle(training_data)
```

[We create 2 arrays now 1 for the image data and another one for labels (with mask or without mask) and start to append the data for the arrays, and after that we did normalization by dividing every image by 255.0 so the image will be uniformly stored]

```python
X = [] #data
y = [] #label
for features,label in training_data:
    X.append(features)
    y.append(label)
X = np.array(X).reshape(-1, IMG_SIZE, IMG_SIZE, 3)
X = X/255.0
Y = np.array(y)
```

[We used pickle to store the data in the pickle format so we can even use this at any time no need to do the whole process again whenever we need to run the program]

The choice of Python stems from its simplicity and consistency, offering a streamlined approach to coding compared to languages requiring extensive lines of code for a singular task. Python facilitates concise and readable code, allowing developers to focus on solving machine learning (ML) problems without being encumbered by technical intricacies. Its simplicity does not compromise the capability to handle complex algorithms and versatile workflows inherent in ML and AI. Python's learnability contributes to its popularity, and its code, being human-readable, enhances model-building in machine learning. Python's intuitiveness surpasses that of other programming languages, a quality appreciated by developers. Furthermore, Python boasts an array of frameworks, libraries, and extensions, simplifying the implementation of diverse functionalities. Recognized as suitable for collaborative implementation, Python, being a general-purpose language, excels in performing intricate machine learning tasks and facilitates rapid prototype development for testing machine learning products. Because implementing AI and ML algorithms can be difficult and time-consuming, Python provides a "Extensive selection of libraries and frameworks" that make working on these projects easier. To encourage developers to write the best code possible, an environment that is well-tested and organized is essential. Many Python frameworks and modules are used by programmers to shorten the development time. Developers can solve typical programming tasks with prewritten code included in software libraries. With its broad library for machine learning and artificial intelligence, Python boasts a strong technology stack. The source code of execution are shown below:

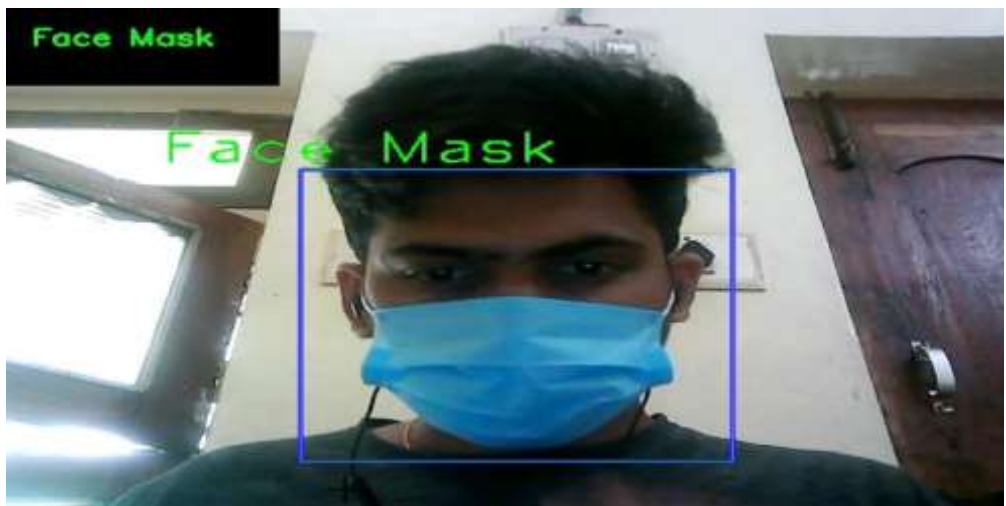

Fig. 4. Detecting Face Mask

The model discerns whether a person is wearing a mask or not, displaying "Face Mask" on the screen when a mask is detected and showcasing "No Mask" if the person is not wearing one.

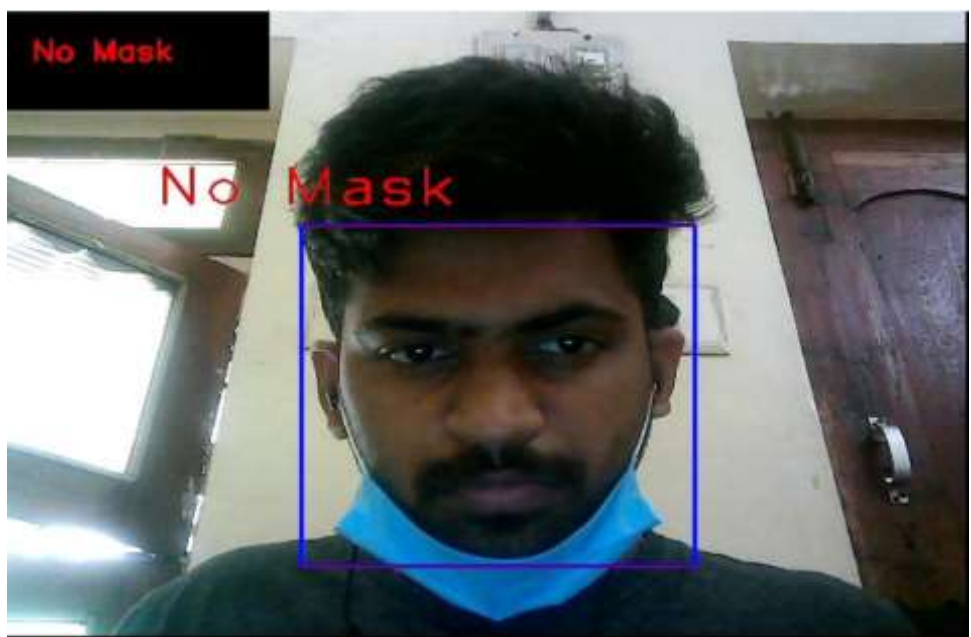

Fig. 5. Detecting No Mask

**5.1. Results Analysis:** The predicted result by the model will be plotted by graphs and tables to check the accuracy and loss precision and after that it will be compared with some other top performing models by using Confusion matrix and comparing the both models performance. Prediction Analysis in the model that is trained using MobileNetv2, the trained model will be saved for comparison. So, when we take a random input image to detect in which the person wearing mask or not. The model trained in that type i.e., if the mask is there on person's face, then the predicted output will be in negative array number. In this image we can easily saw that the output array is in negative value, so every picture with face mask will be predicted as negative value, so even in live demo the photo will be captured frame by frame and went through this process and if the value comes is below zero then it will show output as No Mask.

**5.2 Model Evaluation:** This is the model evaluation table in this table. The output of the trained epochs analysis provides information on the precision with which the model can predict both with and without a mask, as well as the accuracy and precision percentage of the predictions made with and without a mask. It also provides specifics regarding Recall, F1-Score, and Support (image data) to provide a clear understanding of the model's training process. A stable accuracy line indicates that additional iterations are not required to improve the model's accuracy. Making the model assessment as indicated in Table 1 is the next step after that.

Table. 1 Training Loss and Accuracy graph

|  | **Exactness** | **Remember** | **F1 – Score** |
|---|---|---|---|
| Coverage | 0.98 | 0.83 | 0.90 |
| without a face covering | 0.85 | 0.98 | 0.91 |
| Reliability |  |  | 0.91 |
| Average Size | 0.92 | 0.90 | 0.90 |
| Putative Average | 0.92 | 0.91 | 0.90 |

**5.3 Training and Validation of Loss and Accuracy Graphs:** Following the post-processing phase of our model, it yields several models exhibiting frequent instances of high accuracy. Through subsequent processing, a heightened accuracy of 0.98 is achieved, with a validation loss of 0.0855 and validation accuracy of 0.9637. Graph. 1. illustrates the accuracy curve, juxtaposing training accuracy and validation accuracy over various epochs. This visual representation considers two key parameters: training accuracy and validation accuracy. Similarly, it depicts the Loss curve, contrasting training loss and validation loss against the number of epochs. This graphical representation encompasses two crucial parameters: training loss and validation loss.

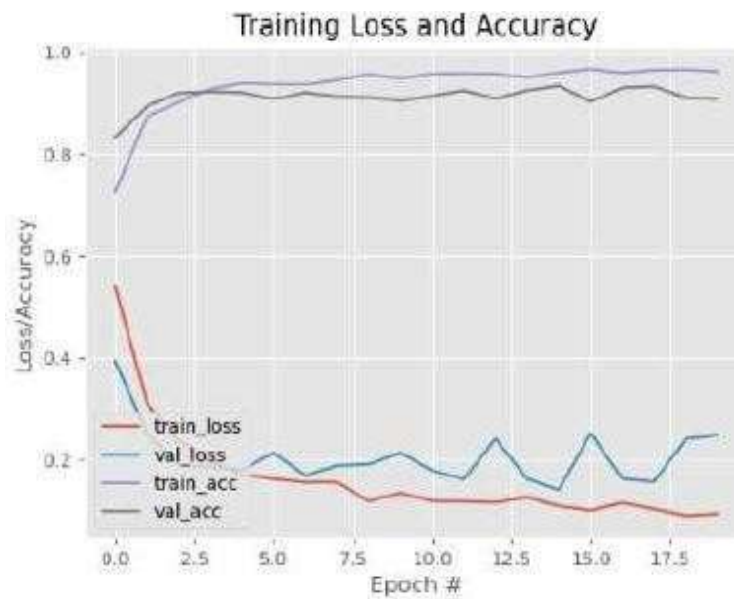

Graph. 1. Accuracy curve

## 6. Conclusion and Future Work

This research collaborates with CNN for the secure detection of face masks, implementing a robust security alert system to enhance surveillance in the designated area. Despite working with relatively small datasets, the system exhibits high accuracy, forming the foundation for the proposed project layer and ensuring favourable outcomes. We advocate for the practical application of this approach in detecting faces with or without masks, contributing potentially to public healthcare. The primary goal of our research is to yield effective results and establish a reliable recognition system. Our future includes exploring advanced feature selection techniques and specialized machine learning algorithms, incorporating larger datasets to tackle more complex challenges. The intention is to enhance our Face Mask Detection tool and release it as an open-source project. This software, compatible with various cameras, can be integrated into web and desktop applications, enabling operators to receive real-time notifications and images in case of individuals without masks. Additionally, an alarm system can be implemented to alert when someone without a mask enters a monitored area. The software can also be linked to entrance gates, permitting only individuals wearing face masks to enter. While our current project may not guarantee face detection from every angle, future development aims to achieve seamless functionality from all perspectives. In the context of the ongoing pandemic, criminal activities, such as theft of oxygen cylinders and concentrators, have increased, often perpetrated by individuals wearing face masks. The proposed model, capable of detecting and recognizing individuals even with masked faces, holds the potential to contribute to reducing such crimes across the country.